\documentclass[manuscript,screen,acmlarge,nonacm]{acmart}
\usepackage{booktabs} 
\usepackage{xcolor}
\usepackage{soul}
\usepackage{changepage}
\usepackage{array}
\usepackage{tabularx}  
\usepackage{seqsplit}  
\usepackage{algorithm}
\usepackage{algpseudocode}
\usepackage{natbib}

\usepackage{enumitem}

\usepackage{pifont}
\newcommand{\cmark}{\ding{51}}%
\newcommand{\xmark}{\ding{55}}%

\setcopyright{none}

\acmDOI{xx.xxx/xxx_x}

\acmJournal{HEALTH}

\begin{document}
\title{Enhancing Performance and User Engagement in Everyday Stress Monitoring: A Context-Aware Active Reinforcement Learning Approach}

\author{Seyed Amir Hossein Aqajari} 
\affiliation{
\institution{University of California, Irvine}
  \city{Irvine} 
  \state{CA} 
  \country{USA}
  \postcode{92617}  
}
\email{saqajari@uci.edu}
\authornote{Both authors contributed equally to this research.}

\author{Ziyu Wang} 
\affiliation{
\institution{University of California, Irvine}
  \city{Irvine} 
  \state{CA} 
  \country{USA}
  \postcode{92617}  
}
\email{ziyuw31@uci.edu}
\authornotemark[1]

\author{Ali Tazarv} 
\affiliation{
\institution{University of California, Irvine}
  \city{Irvine} 
  \state{CA} 
  \country{USA}
  \postcode{92617}  
}
\email{atazarv@uci.edu}

\author{Sina Labbaf} 
\affiliation{
\institution{University of California, Irvine}
  \city{Irvine} 
  \state{CA} 
  \country{USA}
  \postcode{92617}  
}
\email{slabbaf@uci.edu}

\author{Salar Jafarlou} 
\affiliation{
\institution{University of California, Irvine}
  \city{Irvine} 
  \state{CA} 
  \country{USA}
  \postcode{92617}  
}
\email{jafarlos@uci.edu}

\author{Brenda Nguyen} 
\affiliation{
\institution{University of California, Irvine}
  \city{Irvine} 
  \state{CA} 
  \country{USA}
  \postcode{92617}  
}
\email{brendn3@uci.edu}

\author{Nikil Dutt} 
\affiliation{
\institution{University of California, Irvine}
  \city{Irvine} 
  \state{CA} 
  \country{USA}
  \postcode{92617}  
}
\email{dutt@ics.uci.edu}

\author{Marco Levorato} 
\affiliation{
\institution{University of California, Irvine}
  \city{Irvine} 
  \state{CA} 
  \country{USA}
  \postcode{92617}  
}
\email{levorato@uci.edu}

\author{Amir M. Rahmani} 
\affiliation{
\institution{University of California, Irvine}
  \city{Irvine} 
  \state{CA} 
  \country{USA}
  \postcode{92617}  
}
\email{a.rahmani@uci.edu}

\renewcommand{\shortauthors}{Aqajari et al.}

\begin{abstract}
In today's fast-paced world, accurately monitoring stress levels is crucial. Sensor-based stress monitoring systems often need large datasets for training effective models. However, individual-specific models are necessary for personalized and interactive scenarios. Traditional methods like Ecological Momentary Assessments (EMAs) assess stress but struggle with efficient data collection without burdening users. The challenge is to timely send EMAs, especially during stress, balancing monitoring efficiency and user convenience. This paper introduces a novel context-aware active reinforcement learning (RL) algorithm for enhanced stress detection using Photoplethysmography (PPG) data from smartwatches and contextual data from smartphones. Our approach dynamically selects optimal times for deploying EMAs, utilizing the user's immediate context to maximize label accuracy and minimize intrusiveness. Initially, the study was executed in an offline environment to refine the label collection process, aiming to increase accuracy while reducing user burden. Later, we integrated a real-time label collection mechanism, transitioning to an online methodology. This shift resulted in an 11\% improvement in stress detection efficiency. Incorporating contextual data improved model accuracy by 4\%. Personalization studies indicated a 10\% enhancement in AUC-ROC scores, demonstrating better stress level differentiation. This research marks a significant move towards personalized, context-driven real-time stress monitoring methods.
\end{abstract}

%
%

\ccsdesc[500]{Applied computing~Health care information systems}
\ccsdesc[500]{Applied computing~Health informatics}

\keywords{Health Informatics, eHealth, Active Learning, Reinforcement Learning, Personalized Machine Learning}

\maketitle

\definecolor{todo}{RGB}{255,0,0}
\definecolor{aj}{RGB}{128,128,128}
\definecolor{ziyu}{RGB}{139,0,0}

\newcommand{\todo}[1]{\textcolor{todo}{\hl{#1}}}
\newcommand{\aj}[1]{\textcolor{aj}{\hl{#1}}}
\newcommand{\ziyu}[1]{\textcolor{ziyu}{\hl{#1}}}
\newcommand{\thickhline}{\noalign{\hrule height 1pt}}

\section{Introduction}
As per data from the American Institute of Stress \cite{americanstress}, approximately 55\% of individuals in the United States encounter stress throughout their day. The American population stands as one of the most stressed globally, with their current stress levels surpassing the global average by 20 percentage points. The impact of stress extends to the physical body, cognitive processes, emotions, and behavior \cite{mayoclinic}. Unaddressed stress can contribute to several health issues, including high blood pressure, heart ailments, obesity, and diabetes \cite{mayoclinic}. Consequently, daily life monitoring of stress has garnered significant significance within our society, and the advancement of techniques for diagnosing human stress holds paramount importance.

The presence of stress within the human body can be diagnosed by analyzing psychophysiological signals such as Photoplethysmography (PPG) \cite{charlton2018assessing, wang2024differential}. PPG is a simple optical sensing technique for detecting blood volume alterations within peripheral circulation \cite{allen2007photoplethysmography}. With the rapid advancements in technology and the development of the Internet of Things (IoT) \cite{yao2020privacy, wang2020guardhealth, kanduri2023edge, alikhani2023dynafuse}, the acquisition of PPG signals has been greatly facilitated, primarily through the utilization of wearable devices like smart rings or smartwatches \cite{castaneda2018review}. Consequently, monitoring stress levels in everyday life becomes an attainable endeavor by analyzing the PPG signals garnered from the aforementioned wearable devices. Furthermore, the evolution of mobile apps~\cite{cheng2024vetrass, alikhani2024seal, cheng2024efflex} designed for context logging has provided a means to consistently observe and record a user's contextual data, including elements like their location, activities, weather conditions, and other relevant variables, all in real-time \cite{sannino2014mobile, yang2022zebra}. Prior studies have already demonstrated the significance of this contextual information in understanding and identifying stressful experiences encountered by individuals \cite{stojchevska2022assessing, can2020real, han2020objective}. In everyday situations, biosignals vary widely among individuals due to physiological and lifestyle differences, as well as the diverse activities one might partake in. Additionally, the perception of stress levels differs from person to person, leading to biases in the data collected. Therefore, there is a significant need to tailor predictive models to each individual across their various activities. These adjustments, which are essential at the time of deployment, present both conceptual and technical challenges.

The evolution of stress monitoring in daily settings has seen a significant transformation. Originally, stress monitoring was largely confined to controlled environments such as laboratories, which allowed researchers to closely observe and study physiological responses under stress \cite{fahrenberg2007ambulatory, steptoe2003socioeconomic}. Such controlled studies laid the groundwork for understanding stress responses in a structured setting. Over time, advancements in technology enabled the transition to more naturalistic and dynamic environments. This shift paved the way for methods like Ecological Momentary Assessments (EMAs) \cite{burke2017ecological}. EMAs revolutionized stress monitoring by allowing real-time data collection about participants' stress levels throughout their day-to-day activities. 
Using self-reported stress (EMA) as the labeling source, users are asked to respond to real-time queries that link the data gathered by sensors to stress labels in everyday situations. One of the main challenges is optimally collecting these labels (stress levels) from individuals in their daily lives \cite{larradet2020toward}. Frequently triggering EMAs or sending them at inappropriate times, such as when a user is busy with work or sleeping, could burden the user. This could lead to a significantly lower number of reported labels. Moreover, selecting the optimal moments to send the EMAs, especially during instances when an individual is experiencing stress, poses a considerable challenge and holds the utmost importance.



To tackle these challenges, in this work, we introduced a contextual variant of active learning, based on Deep Q-Learning, which incorporates the contextual information pertaining to an individual into the decision-making process. In the initial phase of our research, a context-aware active reinforcement learning algorithm was utilized in an offline setting \cite{tazarv2023active}. This approach was implemented to thoroughly evaluate the effectiveness of our proposed method. We demonstrated that the utilization of such an algorithm in a stress detection task can lead to a reduction of up to 88\% in the required EMAs when compared to a random selection approach, and up to 32\% when compared to traditional active learning methods. Furthermore, we observed that employing such an algorithm can increase the performance of stress detection tasks by up to 21\% compared to a random selection method, and up to 8\% when compared to traditional active learning approaches. However, an offline context-aware active reinforcement learning algorithm abstains from employing active learning to initiate EMAs. However, this approach may still entail user burdens and result in triggering EMAs at inappropriate times. 

In this article, an extension of our previous work \cite{tazarv2023active}, we have improved our proposed algorithm for application in an online setting, leveraging active learning to initiate EMAs. In the online setting, our algorithm initiates EMAs at various points during the study, guided by the contextual information pertaining to the user. Within the context of obtaining participant labels such as stress levels, the active learning algorithm analyzes contextual information in real-time to ascertain the most appropriate timing for posing questions. This adaptive approach serves to reduce participant burden while simultaneously enhancing label accuracy. To comprehensively assess the efficacy of our online algorithm, we compare it to the prior offline one, we conducted two distinct analyses on the same dataset: one employing the offline context-aware active reinforcement learning algorithm and the other utilizing the online variant. Our findings unequivocally demonstrate that the online algorithm yields a substantial enhancement in the performance of the stress detection task when contrasted with its offline counterpart. 
Lastly, we employ a personalization technique to investigate the effects of personalized customization in enhancing the model’s performance. 

In summary, the key contributions of this paper are as follows:

\begin{itemize}[topsep=0pt]
    \item Propose a new form of active learning, utilizing Deep Q-learning, aimed at enhancing interaction with the monitored individual during data collection.
    \item Develop a sensor-edge-cloud layered system architecture for the acquisition and labeling of the data aimed at real-time stress detection. We further demonstrate the effectiveness of our proposed system through the utilization of actual real-time data and comparing it with an identical offline variant of the algorithm.
    \item Incorporate the contextual features into the stress detection models for the purpose of systematically monitoring the influence of contextual factors within the context of stress detection.
    \item Examine how the performance of our algorithm is enhanced with the inclusion of subject-specific data during the training phase in order to explore the influence of personalization on stress detection.
   \item Conduct a two-stage IRB-approved study on 54 individuals across undergraduate and graduate student populations over two periods: June 2020 to June 2021 (offline method) and March 2022 to May 2023 (online method), generating a total of 132,598 filtered samples. We commit to publicly releasing both datasets following our paper's acceptance.

    
\end{itemize}

This paper is structured as follows: Section 2 offers a comprehensive review of stress assessment methods and associated research, highlighting the importance of personalizing models and the crucial role of user behavior and contextual data in real-time labeling. It also emphasizes the need to enhance user engagement in everyday settings. Section 3 details the platform developed for data collection and analysis. Section 4 outlines the dataset we gathered and our data processing methods. Section 5 introduces our proposed context-aware active learning approach, which incorporates user behavior and context into its query mechanism, along with temporal data correlations, to enhance query scheduling. Section 6 reports on the outcomes of various querying techniques in relation to personalization. Finally, Section 7 concludes the paper and includes a discussion.

\section{Related Work}

Stress-related research often delineates its origins from both exogenous factors—such as lifestyle, interpersonal relationships, and financial stability—and endogenous factors like individual psychological constitution and thought processes. These factors act as progenitors for negative affective states, including anxiety and fear, and instigate corresponding physiological responses. The physiological aspect of stress, denoted as a stress response, is a series of bodily reactions to environmental stimuli or stressors. Within the scientific discourse, the construct of stress is categorized into psychological, behavioral, and physiological dimensions. Historically, self-report measures such as the Perceived Stress Scale (PSS), formulated by Cohen et al. \cite{cohen1994perceived}, and the stress inventory by Holmes and Rahe \cite{holmes1967social}, have been the standard for gauging stress levels retrospectively.

Nonetheless, the accuracy of survey-based assessments of stress is compromised by measurement biases, including response bias, which reflects the influence of the query's framing on the participant's responses. Additionally, while some behavioral expressions of stress—like facial expressions—are spontaneous, they may also be subject to volitional control, thus potentially skewing data accuracy. Consequently, recordings of such behaviors must be critically examined for systematic errors that may misrepresent the actual stress magnitude.

Given these constraints, and paralleling the evolution of high-precision sensor technology, there is an augmented demand for veritable detectors of physiological stress markers. Biosignal attributes of stress episodes are typically involuntary, and such data can be acquired through methodologies like electrocardiography (ECG), photoplethysmography (PPG), electromyography (EMG), skin conductance (SC) or electrodermal activity (EDA), respiratory rate (RSP), skin temperature (ST), pupil dilation (PD), and cerebral activity as captured by electroencephalography (EEG) \cite{giannakakis2019review}.



The current methodologies for monitoring stress in daily life utilize EMAs to inquire about participants’ stress levels throughout the day \cite{larradet2020toward}. 
The task of effectively gathering accurate stress level indicators from individuals in the context of their everyday activities presents a significant challenge \cite{settles2009active}. The over-frequent activation of EMAs or their issuance at times that clash with a user's schedule, such as during work hours or rest periods, can be an imposition. This may culminate in a reduced quantity of reported stress labels. Additionally, pinpointing the precise moments for sending EMAs, particularly in moments when stress levels are elevated, is of paramount significance and presents a notable challenge.

Existing methods in the literature for the deployment of EMAs in daily life stress monitoring studies can be categorized into three distinct categories: 1. Random, 2. Time-based, and 3. Statistical-based.

Within the random triggering methods, EMAs are dispatched at random intervals throughout the course of the study. Random sampling is often the preferred approach in situations where the research topic's indicators cannot be reliably ascertained \cite{dogan2022stress}. However, when the research topic possesses specific objectives and focus, alternative methods tend to yield more favorable outcomes.

Time-based triggering methods involve sending EMAs at fixed pre-defined intervals throughout the day. The majority of existing research endeavors in daily life stress monitoring in the literature employ this algorithm for their label querying system, as documented in previous studies \cite{yu2022semi,yu2020passive,mundnich2020tiles,wang2020social,battalio2021sense2stop}. While this algorithm boasts simplicity of implementation and uniform coverage of the study period, it imposes a considerable burden on participants due to untimely EMA deliveries. Consequently, this may result in an increased prevalence of missing data and a reduction in the utility of collected labels.

In the context of statistical-based triggering methods, EMAs are dispatched based on the distribution of samples \cite{tazarv2021personalized}. Under this triggering algorithm, a label is requested for a specific sample based on the number of unlabeled samples within its vicinity. Although this policy can effectively mitigate the incidence of undesired EMA deliveries, it may still impose a burden on users and lead to missing data, as it fails to consider contextual information about users, which is pivotal in determining the optimal times for EMA deployment.

In this study, we propose a context-aware active reinforcement learning approach to effectively trigger EMAs throughout the day.

Initially we conducted an offline study where we employed a statistical-based triggering method to send EMAs throughout the day. In this phase, the probability of selecting each sample for labeling is proportionate to the quantity of prior unlabeled samples in its proximity. This approach increases the likelihood of requesting a user label for a sample situated in a region with a substantial number of unlabeled samples. Upon accumulating a sufficient number of labeled samples for each region, the data collection process is terminated. We implemented three distinct algorithms offline for optimal label selection in model development: 1. Random, 2. Traditional Active Reinforcement Learning, and our novel approach 3. Context-Aware Active Reinforcement Learning. Our findings revealed that using a context-aware active reinforcement learning algorithm in stress detection significantly decreases the necessity for EMAs enhances the effectiveness of stress detection over random or traditional active learning methodologies. As previously noted, statistical-based triggering algorithm may still impose a user burden and result in missing data due to its failure to incorporate user contextual information into the label-querying decision-making process.

In the next phase, we propose an online context-aware active reinforcement learning algorithm to utilize RL agent for decision making in real time to further improve the performance. This algorithm actively utilizes a context-aware active learning approach in real-time based on Deep Q-Learning to determine whether an EMA should be triggered for a given sample. By considering real-time contextual information related to each user in the decision-making process of whether to trigger an EMA for a specific sample, our approach is poised to significantly reduce the user burden associated with untimely EMA deliveries, consequently leading to an increase in the acquisition of high-utility labels. Table \ref{related-works} provides a summary of the comparison between our work (offline and online studies) and existing literature on this subject.

\begin{table*}[!ht]
\centering
\caption{{\bf Comparison of our study vs existing works}}
\vspace{-3mm}
\begin{tabularx}{\textwidth}{|X|X|X|X|X|X|X|}
\hline
\textbf{Study} & \textbf{Triggering Method} & \textbf{EMA Frequency} & \textbf{Real-time Analysis} & \textbf{Data-based queries} & \textbf{Context-based queries} & \textbf{Online triggering method}\\ \hline
Yu et al. \cite{yu2022semi} & Time-based & 1.5 hours apart, 10/day & \xmark & \xmark & \xmark & \xmark \\
\hline
Yu et al. \cite{yu2020passive} & Time-based & 1/day & \xmark & \xmark & \xmark & \xmark \\
\hline
Mundnich et al. \cite{mundnich2020tiles} & Time-based & 1/day & \xmark & \xmark & \xmark & \xmark \\
\hline
Wang et al. \cite{wang2020social} & Time-based & Every three months & \xmark & \xmark & \xmark & \xmark \\
\hline
Battalio et al. \cite{battalio2021sense2stop} & Random and Time-based & End of the day, varies & \cmark & \cmark & \xmark & \xmark \\
\hline
Our Offline Study & Statistical-based & A cap of seven EMAs per day & \cmark & \cmark & \xmark & \xmark \\
\hline
\textbf{Our Online Study} & \textbf{Active Reinforcement Learning} & \textbf{A cap of seven EMAs per day} & \cmark & \cmark & \cmark & \cmark \\ \hline
\end{tabularx}
\label{related-works}
\end{table*}

\section{Offline Study}

\begin{figure}[t]
      \includegraphics[width=0.7\linewidth]{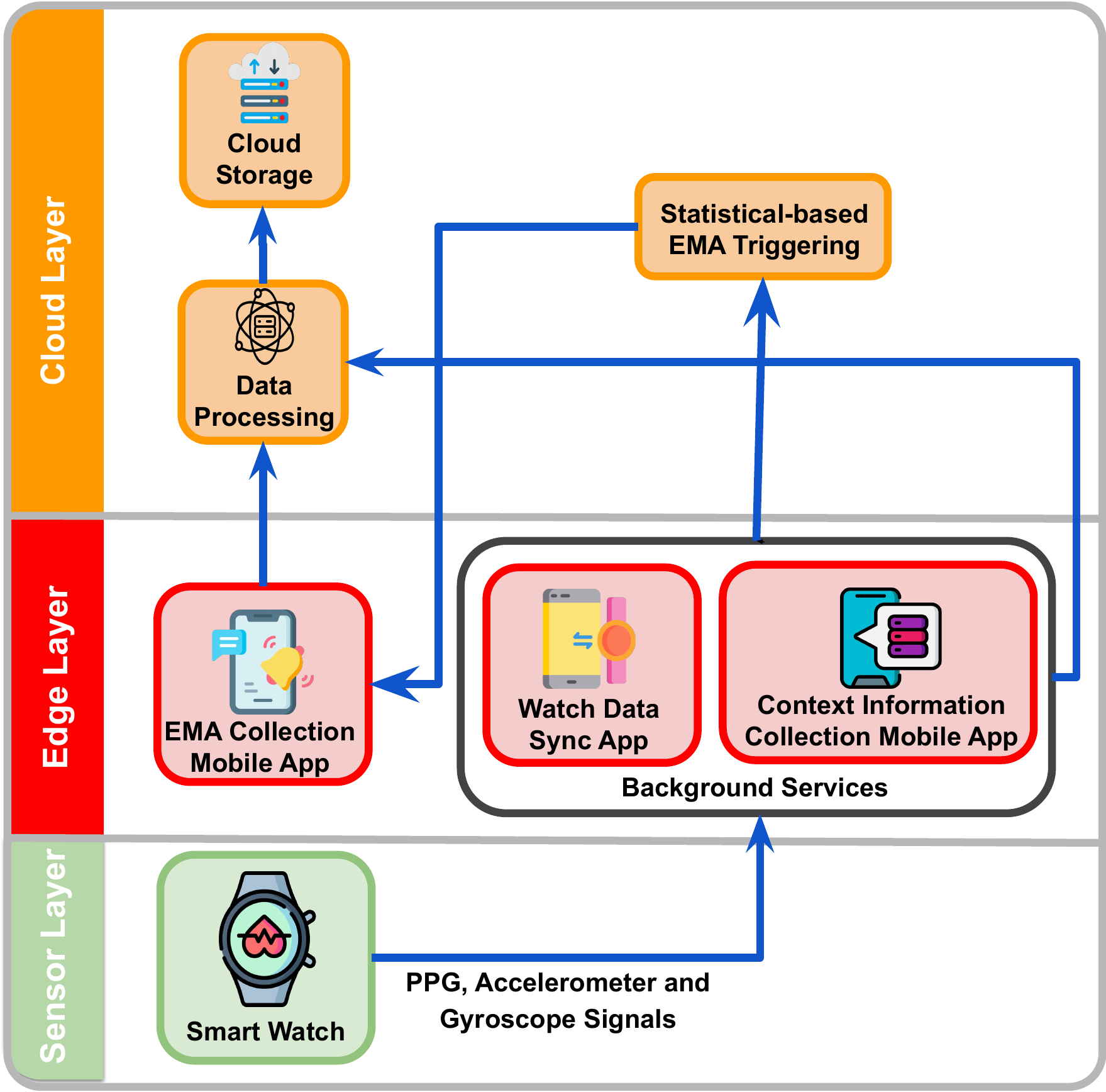}
      \vspace{-2mm}
      \caption{System Architecture - Offline Study}
      \label{fig:arch_offline}
      \centering
      \vspace{-6mm}
\end{figure}


In the initial stage of this work, we target to evaluate the effectiveness of our proposed label triggering method \cite{tazarv2023active}. The EMAs are dispatched to participants' phones on a statistical-based basis. Once data collection was completed, we applied our proposed method of context-aware active reinforcement learning for the labeling process. Our offline study involved an Institutional Review Board (IRB)-sanctioned study on human subjects, during which we gathered over 2,629 days of data in everyday environments from college students. 

The collected dataset encompasses PPG and various motion metrics (such as acceleration, gyroscope, and gravity readings), and is partially annotated with information on stress levels, emotional states, and physical activities, determined through EMAs conducted at statistical-decided intervals.

Our data collection initiative, spanning from June 2020 to June 2021, involved 20 volunteers selected from undergraduate and graduate student populations. The demographic breakdown of the participants included 13 male and 7 female students, ranging in age from 19 to 29 years. During the study, we gathered 109,586 samples over a period of 2,629 days. Participants contributed to the study for periods ranging from 11 to 287 days, with an average participation duration of 131 days. On average, each subject contributed 5,479 instances to the dataset. We commit to making this dataset publicly available following the acceptance of our paper. 

\subsection{Proposed System Architecture}

Creating a reliable system to gather real-time physiological and contextual data while using active learning from participants is challenging. Wearable devices like smartwatches can be affected by motion artifacts, requiring extensive processing for stress detection \cite{seok2021motion}. Timing label requests is crucial to ensure participant engagement and label reliability. Figure \ref{fig:arch_offline} depicts our offline proposed three-layer system including a sensor layer for data acquisition, an edge layer for data transmission and user interaction, and a cloud layer for data processing and decision-making respectively. 
This system architecture illustrates the architectural composition of our proposed three-layer system, called ZotCare \cite{zotcare}.


\subsection{Sensor layer}

This study utilizes Samsung Galaxy Active 2 Watches, equipped with PPG (20Hz), accelerometer, and gyroscope sensors \cite{sarhaddi2022comprehensive}. We developed a Tizen-based smartwatch app to collect these signals \cite{vashisht2014study}. Data is sent to the cloud via Wi-Fi or Bluetooth to a smartphone when Wi-Fi is unavailable. The raw signal acquisition program consists of two services and a user interface (UI). The first service sends sensor data to the cloud every 15 minutes at 2-minute intervals.

\subsection{Edge layer}

We employ the AWARE framework \cite{awareframework} to collect contextual data in everyday scenarios. AWARE is an open-source mobile tool designed for recording, sharing, and reusing context-related information on mobile devices. It utilizes the built-in sensors of smartphones to capture various aspects of daily life, including battery status, weather conditions, location, screen activity, and more. In situations where Wi-Fi connectivity is unavailable, we utilize an alternative smartphone application installed on the edge of our network. This application collects raw PPG signals and accelerometer data from the sensor layer through Bluetooth and subsequently transmits this data to cloud storage. To obtain stress level ratings from our study participants, we have developed an additional smartphone application. This application employs an EMA approach to request stress level assessments from the participants.

\subsection{Cloud layer}

This layer comprises two distinct modules:

\begin{itemize}
    \item \textbf{Data Processing:} This module focuses on processing data retrieved from the edge layer, with its primary objectives being encryption and the storage of data in the appropriate format on ZotCare servers.
    \item \textbf{Statistical-based EMA Triggering:} Our label triggering method is consists of two phases:

\begin{itemize}

\item \textbf{Initial Stage:} To obtain an initial approximation of the sample distribution within the sample space, we start the procedure with observation. During the first N samples (100 samples in our configuration, equivalent to approximately 25 hours of wearing the watch), no EMAs are initiated. By the conclusion of this phase, an estimation of the sample distribution in the sample space is obtained.

\item \textbf{Query Stage:} Subsequent to the initial phase (from N+1 onward), EMAs are triggered (labels requested) for a subset of samples. The selection probability for labeling each sample is proportionate to the number of preceding samples (unlabeled) in its proximity. This approach ensures that samples in regions with a substantial number of unlabeled counterparts are more likely to be queried for labels. Once a sufficient number of labeled samples are acquired for a particular region, the label collection for that region ceases. Nonetheless, the minimum probability of triggering an EMA for a sample is set at P = 0.1. Consequently, even if a sample is situated in an area with few or no previous samples, the probability of a query remains nonzero. This design enables exploration of unseen regions as well as regions with higher densities while maintaining a balanced approach.

\end{itemize}
\end{itemize}

\subsection{Preprocessing}
\label{ft_ppg}

Following the conclusion of study and data collection, the collected raw PPG signals are preprocessed in order to extract relevant features for model building.

\subsubsection{Data Cleaning}
The bio-signals from wearable devices, which are inherently noisy, are stored directly in the cloud. The goal of this step is to remove clearly erroneous data points. We use the motion data to remove noises and artifacts in the bio-signals. In our study, we primarily focus on refining PPG signals and heart-rate data. For PPG signals, we utilize a bandpass Butterworth filter with cutoff frequencies between 0.7 Hz and 3.5 Hz. To ensure consistency, the filter is of the third order, and we adopt a sampling rate of 20 Hz, which aligns with our data collection parameters. Additionally, we implement a moving average over a 1-second window to smoothen the PPG data, mitigating artifacts commonly induced by body gestures and movements in daily environments.

\subsubsection{Data Normalization}

Normalization is indispensable when aiming to minimize variances specific to individual participants and countering the repercussions of subpar bio-signal samples. A standout method in this context, particularly in statistical analyses and machine learning, is the min-max normalization. This technique scales feature values consistently within a predetermined range, commonly [0, 1]. By doing so, it ensures the inherent structure of the data remains intact, which becomes crucial for algorithms that might be affected by the magnitude of the features. For our PPG bio-signals, we've employed the min-max normalization as our primary estimator to ensure that each feature aligns appropriately within a range dictated by the training set.

\subsubsection{Feature Extraction}

In our investigation, a feature extraction module was employed to analyze PPG data in 2-minute intervals. This facilitated the identification of PPG peaks and the derivation of key metrics such as heart rate. Utilizing the HeartPy library \cite{van2019heartpy} for comprehensive PPG signal processing, we extracted 12 features from both electrical activations and pressure waveforms in the dataset. The extracted PPG features are presented in Table \ref{ppg-features}.

\subsection{Context-aware Active Reinforcement Learning Algorithm}
\label{contextaware}

Our study utilizes the Context-aware Active Reinforcement Learning algorithm to label the collected data in our offline study. In this section, we provide a detailed explanation of this method and compare it with the random selection method and traditional active reinforcement learning methods for label querying.

In traditional supervised learning, the entire labeled dataset was utilized for training \cite{kotsiantis2007supervised}. However, within our personalized data collection approach, we accumulated labeled data from diverse sources over time. To leverage this, we iteratively queried our users for new annotations. Commencing with a subset of labeled data, we employed a query mechanism to identify the most informative unlabeled instances with the assistance of human experts.

Active learning played a pivotal role in this process, strategically selecting data samples for labeling to enhance model accuracy while minimizing data usage. Strategies encompassed uncertainty sampling (opting for ambiguous data) and diversity sampling (choosing unique and indicative data). Nevertheless, real-world queries incurred costs, necessitating a delicate balance between query expenses and model improvement.

In our scenario, we encountered distinctive challenges. User behavior influenced data quality and availability, contingent on factors such as activity, time, query frequency, and phone interaction. A lack of response resulted in the denial of labeling and delayed responses, diminishing data quality alignment. Consequently, our active learning approach needed to consider not only data quality but also future label accessibility. We proposed the use of Deep Q-Learning, where an agent modeled user behavior to ensure sustained user engagement.

\subsubsection{Deep Q-learning}
Deep Q-Learning (DQN) \cite{hester2018deep} is a model-free, online, off-policy reinforcement learning method. At its core, DQN seeks to estimate the action-value function, denoted as \( Q(s, a) \), which predicts the expected return after taking an action \( a \) in state \( s \). The Bellman equation, which is fundamental to Q-learning, is given by:
\begin{equation}
Q(s, a) = r + \gamma \max_{a'} Q(s', a')
\label{eq:bellman}
\end{equation}
Where \( r \) is the immediate reward, \( \gamma \) is the discount factor, and \( s' \) is the subsequent state after taking action \( a \) in state \( s \). The primary distinction between traditional Q-learning and DQN is the utilization of deep neural networks to approximate the Q-values. This is paramount for tasks with large state spaces. The loss \( \mathcal{L} \) during training is defined as:
\begin{equation}
\mathcal{L}(\theta) = \mathbb{E}_{(s, a, r, s') \sim U(D)} \left[ \left( r + \gamma \max_{a'} Q(s', a'; \theta^-) - Q(s, a; \theta) \right)^2 \right]
\label{eq:loss}
\end{equation}
Where \( D \) is the replay buffer, \( U(D) \) is a uniform random sample from \( D \), \( \theta \) are the network parameters, and \( \theta^- \) are the target network parameters. 

In the following part, we will elaborate the detailed definitions in our DQN.

\subsubsection{State Representation}

The state vector, denoted by \( s \), encodes crucial information leading to decision-making. The intention is to refine and personalize the stress detector to optimize accuracy while minimizing user queries. The state comprises:

\begin{itemize}
    \item \textbf{Uncertainty Factor}: Originating from the raw output of a pre-trained classifier. This factor measures the distance from the decision boundary, effectively quantifying the confidence of the prediction.
    \item \textbf{Time-aware Response Rate}: This accounts for the time of the day (in hourly intervals) and embodies the user's responsiveness across different hours.
    \item \textbf{Time since Last Query}: To enhance user experience and prevent excessive querying in short time frames.
    \item \textbf{Time of Day}: Represents the current hour and is used to model potential variations in stress levels throughout the day.
\end{itemize}

\subsubsection{Reward Formulation}

The reward function integrates components from the `n\_state` vector for holistic decision-making. It is formulated as:

\begin{align}
r_0 &= \frac{1}{1 + e^{-20(n_{\text{state}[0]}-0.5)}} \\
r_1 &= \text{reward\_F}(n_{\text{state}[1]}) \\
r_2 &= \frac{1}{1 + e^{-10(n_{\text{state}[2]}-0.5)}}
\end{align}

The overall reward function, \( R \), based on the action taken, is:

\begin{equation}
R(\text{action}) =
\begin{cases}
\text{reward\_p} & \text{if action is True} \\
3 - \text{reward\_p} & \text{if action is False}
\end{cases}
\end{equation}

\subsubsection{Q-Network Design}

The Q-network constitutes the backbone of our framework. The structure and features are enumerated below:

\begin{itemize}
    \item The core is a densely connected neural network geared towards estimating Q-values.
    \item Input: The network takes in 4 nodes, matching the count of state variables.
    \item Hidden Layers: The architecture consists of variable hidden layers, as specified by the list \( h \). In the provided example, four hidden layers are employed with 5, 9, 7, and 5 nodes, respectively. Each of these nodes uses the ReLU activation function and incorporates both \( l1 \) and \( l2 \) kernel regularizers, with the \( l2 \) regularization strength set at \(1e^{-2}\).
    \item Output: The network furnishes 2 output nodes, indicative of the duo of feasible actions, with a linear activation function.
\end{itemize}

Additionally, in our experiments, the agent employed an \(\epsilon\)-greedy policy accompanied by a linear annealing schedule for its exploration factor. This strategy ensures a gradual transition from exploration to exploitation during the learning process, thereby enhancing convergence and robustness in diverse environments. For our implementation, we leveraged the Keras-RL library \cite{plappert2016kerasrl}. To elucidate, the \(\epsilon\)-greedy policy in reinforcement learning can be characterized as follows:

\begin{equation}
\pi(a|s) = 
\begin{cases} 
\epsilon + \frac{1-\epsilon}{|A|} & \text{if } a = \text{argmax}_{a' \in A} Q(s, a') \\
\frac{1-\epsilon}{|A|} & \text{otherwise}
\end{cases}
\end{equation}

Where:
\begin{itemize}
    \item \( \pi(a|s) \) is the probability of taking action \( a \) in state \( s \).
    \item \( \epsilon \) is the exploration probability.
    \item \( A \) is the set of possible actions.
    \item \( Q(s, a') \) is the estimated value of taking action \( a' \) in state \( s \).
\end{itemize}

For our experiment, we employed a sequential memory architecture with a capacity limited to 50,000 instances and a window length set at one. The DQN agent was initialized with parameters set as follows: a discount factor \( \gamma \) at 0.95, a warm-up phase consisting of 100 steps, and a learning rate of \(1e^{-2}\) for the target model update. Optimization was carried out using the Adam optimizer, and the performance was gauged using the Mean Absolute Error (MAE) metric.

\subsubsection{Policy Strategy}

The decision strategy, symbolized by \( \pi_\theta(s, a) \), selects the action that corresponds to the optimum Q-value for a specified state \( s \). To guarantee a complete traversal of the state space:

\begin{itemize}
    \item On an estimated 5\% of occasions, a query is initiated randomly. This procedure considers the potentially undiscovered areas within the state space.
    \item Following every \( K = 100 \) occasions, which ideally occurs once in a 24-hour span, the user's response frequency metrics are re-calibrated according to their interaction patterns. This adjustment acknowledges individual variability and revitalizes the query selection methodology tailored for each user.
    \item The stress detection module undergoes periodic retraining. This process assimilates the most recent subjectively labeled information, amalgamated with the prior objective data procured from diverse subjects.
\end{itemize}

\subsection{Evaluation and Results} \label{offline_result}

We initiated the pre-training of the stress detector, which was subsequently utilized to derive the state and reward in constructing the active learning framework. A random forest classifier with n = 500 estimators (number of trees) and max depth = 5 for each tree was employed. Participants were requested to assess their stress levels on a five-point scale: (1) not at all, (2) a little bit, (3) some, (4) a lot, and (5) extremely. We translated the stress labels into two categories: a lot and extremely as 1 (stressed), and the remaining three labels as 0 (not stressed).

We conducted training for the model using labeled data from 14 different subjects, reserving one subject for personalization. The newly trained model on this subject exhibited a recall value of recall = 0.238 for the minority class (class stressed). The Q-learning agent underwent pre-training with an offline sequential objective dataset. Upon completing a sequence (referred to as one episode), we restarted from the beginning until reaching the total number of steps. The model underwent training for K=200,000 steps. The total reward achieved during the episodes reached a saturation point before this stage, indicating the model's convergence.


The agent underwent training in two distinct modes. Initially, we employed a "traditional" approach, where the agent's attention was solely on the state and reward associated with the classifier's raw output. The agent received significant rewards for executing the 'submit query' action for instances within the classifier's uncertainty region, while being rewarded for the 'do not submit query' action for other instances. In this mode, the agent operated without capturing contextual information and functioned akin to a conventional active learning selection policy. Subsequently, a modification was introduced by incorporating contextual information into the reward function, as previously described. High rewards were assigned for the 'submit query' action for instances not only within the uncertainty region but also from a time interval when the user demonstrated increased responsiveness. Moreover, instances not in a short time distance from the preceding 'select' action were considered. For other instances, the agent received rewards for executing the 'do not submit query' action.

\begin{figure*}[t]
      \includegraphics[width=0.8\linewidth]{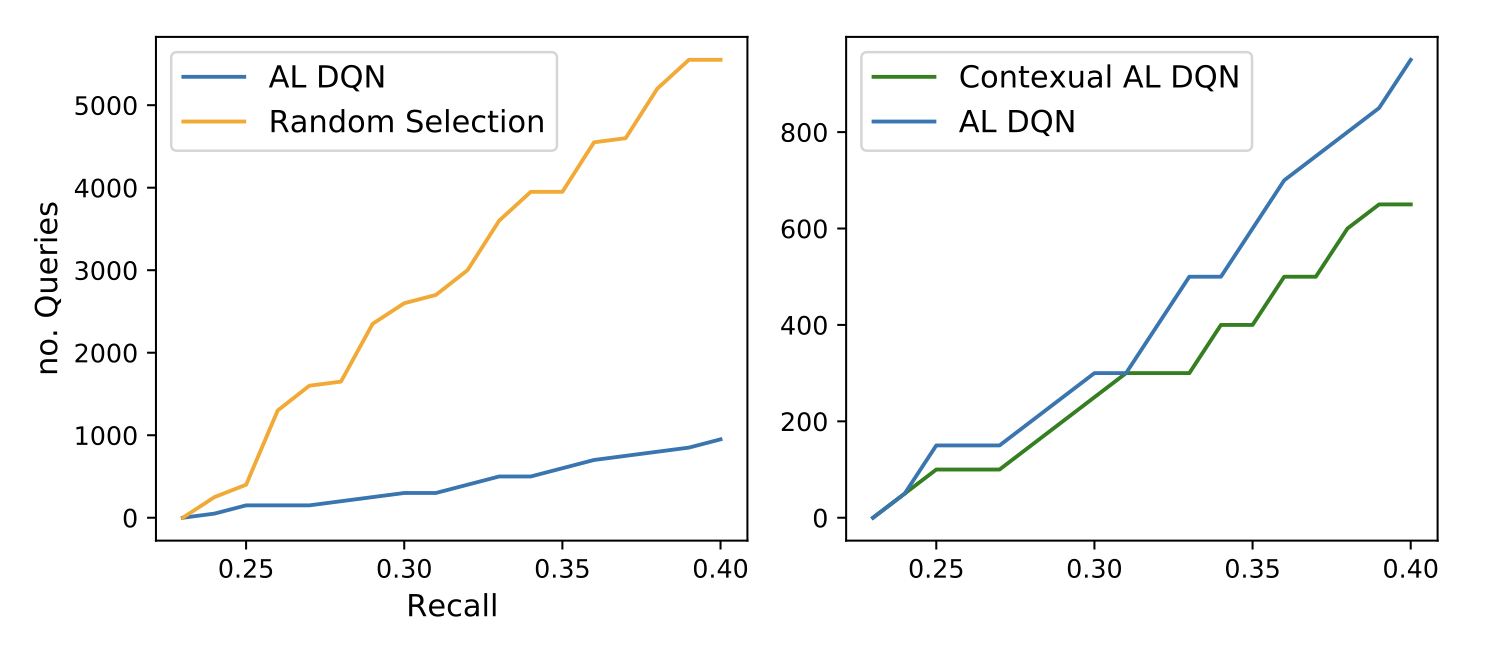}
      \vspace{-3mm}
      \caption{Number of queries needed to reach a certain performance level during personalization.}
      \label{fig:rr}
      \centering
      \vspace{-4mm}
\end{figure*}

We conducted an analysis on the quantity of queries needed to achieve a particular level of personalization. The experiment was repeated N = 100 times to mitigate the influence of random selection. The average number of instances is presented in Figure \ref{fig:rr}. Notably, a substantial disparity exists between random selection and DQN agents, with the context-aware agent demonstrating the capability to attain high performance with a significantly lower number of queries. Specifically, the context-aware selection policy reduces the required queries by up to 88\%, in contrast to the random selection method. While the number of necessary labels remains consistent for the two DQN agents throughout the analysis (as expected), the context-aware agent manages to reduce the required queries by up to 32\%. These findings, derived from a subject with a higher number of labels, exhibit similar trends when extended to data from other subjects.

\begin{figure*}[t]
      \includegraphics[width=0.65\linewidth]{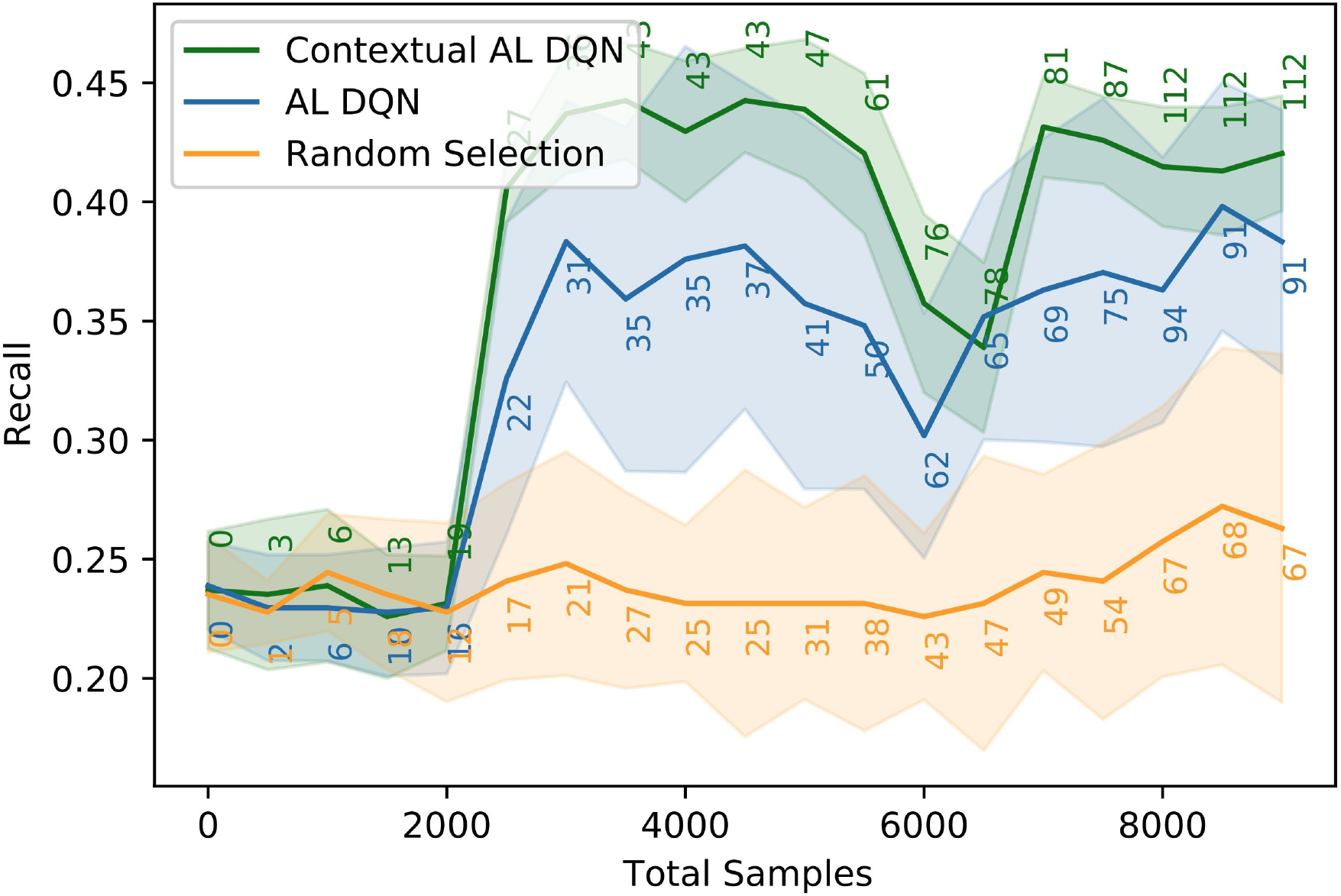}
      \vspace{-3mm}
      \caption{Presonalization Recall in Previous Work}
      \label{fig:pers_previous}
      \centering
      \vspace{-4mm}
\end{figure*}

We examined the effectiveness of two agents and a random selection policy in achieving the primary objective of personalizing the classifier, and the findings are illustrated in Figure \ref{fig:pers_previous}. The number of instances selected for querying remained consistent across each step for all selection methods. However, a subset of these selected instances stayed unlabeled, resulting in different quantities of instances available for personalization depending on the selection policies. Aside from this, the selected instances had varying impacts on personalization under different policies, comparing random selection to the other two methods. From a single subject, we obtained a total of 12,700 instances, with 922 labeled. We reserved 230 labeled instances (from the end of the sequence) as test data from one subject, leaving the remainder for training (25\% - 75\%). The process started without any personalization, gradually progressing through partially labeled subjective data. A subset of this data was chosen for querying, and a part of the selected data was labeled, with the labeled data being utilized for personalization. At each step, the context-aware agent selected fewer instances than the non-context-aware agent. To ensure a fair comparison, we randomly down-sampled the number of queries from the larger group. Additionally, for random selection, we randomly picked a number of partially labeled samples equivalent to the number of queries from the agents. To mitigate the impact of random selection, at each step, we selected instances and personalized the models N = 100 times. Figure \ref{fig:pers_previous} displays the mean and standard deviation of recall (True Positive Rate) for the stressed class on test data for the three selection methods.

Instances that are selected by the random agent do not improve the performance significantly since they include samples from the entire region of the input space of the classifier, including samples whose class is ‘trivial’ to be extracted. Instances that are selected by a non-contextual active learning method (blue curve) increase the performance. However, with an equal number of queries, the best result is achieved when the agent is context aware (green curve), since it results in a higher number of impactful instances which also have a higher chance to receive the label.

\section{Online Study}

In our online study, we employ the Context-aware Active Learning Deep Q-Network (Context-aware AL DQN) algorithm, aligned with our offline investigation, to assess the efficacy of our proposed algorithm in a real-time setting where users are actively involved in training the Reinforcement Learning (RL) agent for decision-making. This real-time approach significantly reduces the user burden and has the potential to enhance stress detection performance compared to its offline counterpart by leveraging a real-time smart RL agent that query labels.

We assessed data derived from a cohort of 34 individuals. This study spanned from March 2022 to May 2023. Participants, ranging in age from 19 to 29 years, provided a comprehensive dataset. After filtering out anomalous and noisy records, we aggregated 23,012 samples over a period of 420 days. On an average basis, each participant yielded 676 distinct samples. It is noteworthy to mention that the respective IRB granted approval for all aspects of this investigation. We pledge to release the dataset to the public once our paper is accepted.

\begin{figure}[t]
      \includegraphics[width=0.7\linewidth]{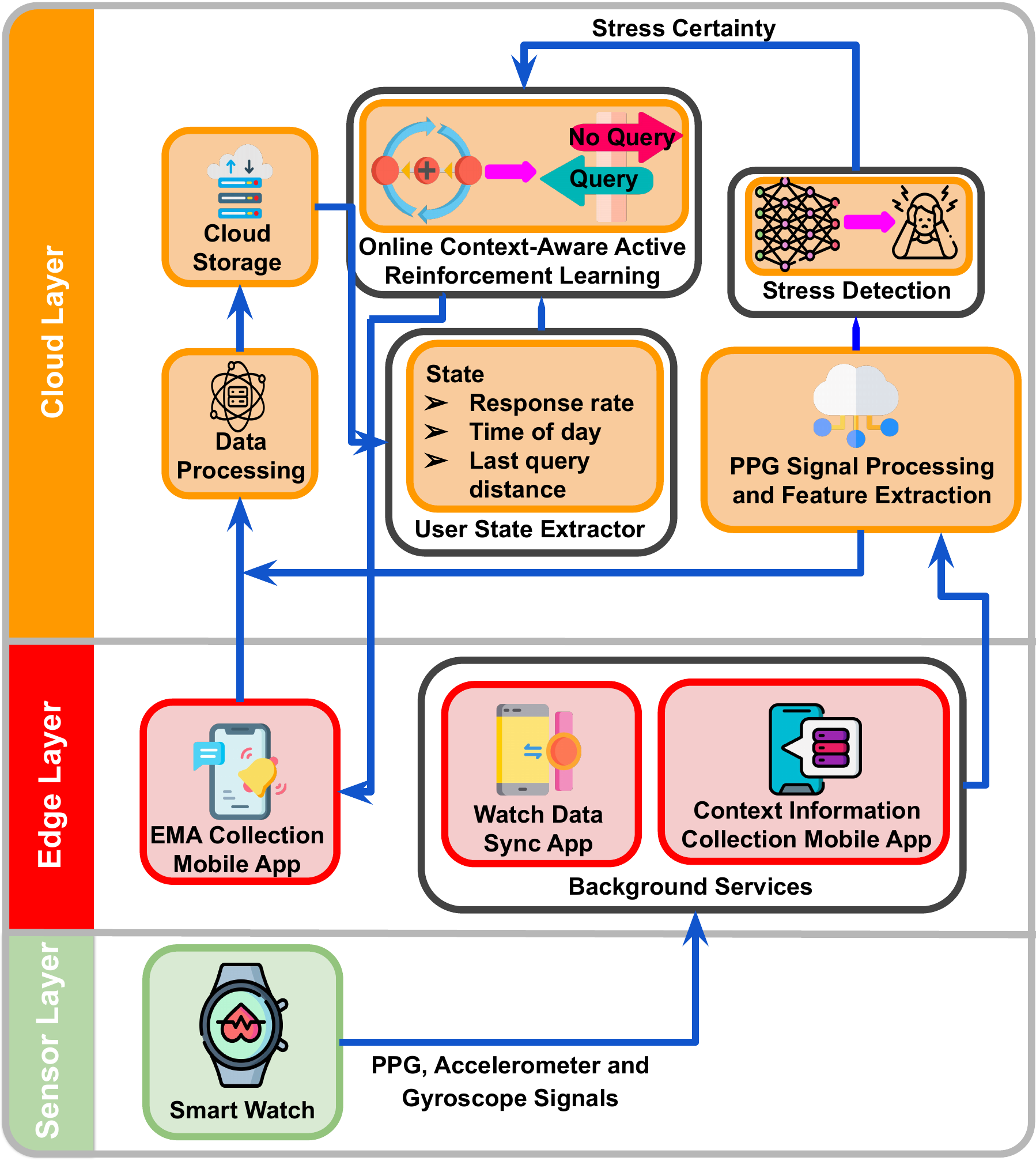}
      \vspace{-2mm}
      \caption{System Architecture - Online Study}
      \label{fig:arch_online}
      \centering
      \vspace{-6mm}
\end{figure}

\subsection{Proposed System Architecture}

The proposed system architecture is illustrated in Figure \ref{fig:arch_online}. Consistent with our offline study, we maintain a three-layer system, denoted as ZotCare. A comparison with the architecture presented in Figure \ref{fig:arch_offline} reveals that, although the sensor layer and the edge layer remain unchanged, substantial modifications are implemented in the cloud layer. These alterations are undertaken to render the system conducive to real-time label querying through the utilization of our proposed context-aware active reinforcement learning algorithm explained in \ref{contextaware}. 

Cloud layer mainly comprises of four distinct modules to replace the previous simple statistical-based triggering method with our proposed triggering algorithm.

\begin{itemize}
    
\item PPG Signal Preprocessing and Feature Extraction: To effectively identify moments conducive to experiencing stress, we continuously monitor participants' stress levels in real time using PPG signals from their watches. However, to make these signals suitable for stress prediction, several preprocessing steps are required. This module is dedicated to preparing the PPG signals for stress detection. More details can be found in Section \ref{preprocessing}.


\item Stress Detection: We utilize the data from our previous study \cite{tazarv2023active} to construct our stress detection module. The features extracted from the PPG signals are input into this module for stress detection. The level of certainty regarding stress is then forwarded to the context-aware active reinforcement learning module to aid in identifying stressful moments.

\item Context Recognition: Within this module, we extract contextual information pertaining to each user, which is subsequently provided to our active learning module for decision-making purposes. This includes factors such as the time elapsed since the last query, the time of day, and the time-aware response rate. The time-aware response rate considers the user's responsiveness within the current hour based on their historical activity.

\item Context-Aware Active Reinforcement Learning: The primary objective of this module is to determine whether it is appropriate to trigger an EMA at any given moment. Stress certainty, time elapsed since the last query, time of day, and time-aware response rate are all input into this module to inform the decision-making process. If an EMA needs to be triggered, a notification is dispatched to the user's mobile device on the edge layer, prompting them to rate their current stress level. In the following section, we will delve deeper into the training process of the active reinforcement learning agent.
\end{itemize}

\subsection{Preprocessing} \label{preprocessing}


PPG signals, contextual AWARE data, and user-reported stress levels are collected from the cloud for stress model construction. However, raw cloud-stored PPG and AWARE data need preprocessing before building the model. This section explains our data preparation steps.

\subsubsection{Data Cleaning and Normalization}

This study employs the same modules for data cleaning and normalization as discussed in the offline study section (see Section \ref{ft_ppg}).


\subsubsection{Feature Extraction}

\begin{itemize}
\item PPG Features: This module has been previously discussed in \ref{ft_ppg}. For the information regarding the PPG features, please refer to the Table \ref{ppg-features}.


\item Contextual Features: The raw contextual information obtained from AWARE is not ready for building the stress detection models. We transform both categorical and numerical raw features into solely numerical features. We show the features extracted from raw AWARE data in Table \ref{aware-features}.
\end{itemize}

\begin{table}[!ht]
\centering
\caption{{\bf AWARE Features}}
\vspace{-3mm}
\begin{tabular}{|l|l|}
\hline
\textbf{Feature} & \textbf{Definition} \\ \hline
Call & Call duration, type, and count\\ \hline
Notification & APP source and count\\ \hline
Screen \& Touch & User screen interactions\\ \hline
Battery & Battery charge duration and level \\ \hline
Message & Message type and count \\ \hline
Time & Time of the day (24-hour format) \\ \hline
Location & Longitude, latitude, altitude\\ \hline
\end{tabular}
\vspace{-5mm}
\label{aware-features}
\end{table}

\begin{table*}[!ht]
\centering
\caption{{\bf PPG Features}}
\vspace{-3mm}
\begin{tabular}{|l|l|}
\hline
\textbf{Feature} & \textbf{Definition} \\ \hline
BPM & Heart beats per minutet\\ \hline
IBI & Inter-Beat Interval, the average time interval between two successive heartbeats (NN intervals)\\ \hline
SDNN & Standard deviation of NN intervals\\ \hline
SDSD & Standard deviation of successive differences between adjacent NNs\\ \hline
RMSSD & Root mean square of successive differences between the adjacent NNs \\ \hline
PNN20 & The proportion of successive NNs greater than 20ms \\ \hline
PNN50 & The proportion of successive NNs greater than 50ms\\ \hline
HR\_mad & Median absolute deviation of NN intervals\\ \hline
SD1 and SD2 & Standard deviations of the corresponding Poincare plot\\ \hline
S & Area of ellipse described by SD1 and SD2\\ \hline
BR & The number of breaths per minute (breathing rate)\\ \hline
\end{tabular}
\vspace{-3mm}
\label{ppg-features}
\end{table*}

\subsubsection{Data Labeling}
\begin{figure}[t]
      \includegraphics[width=0.75\linewidth]{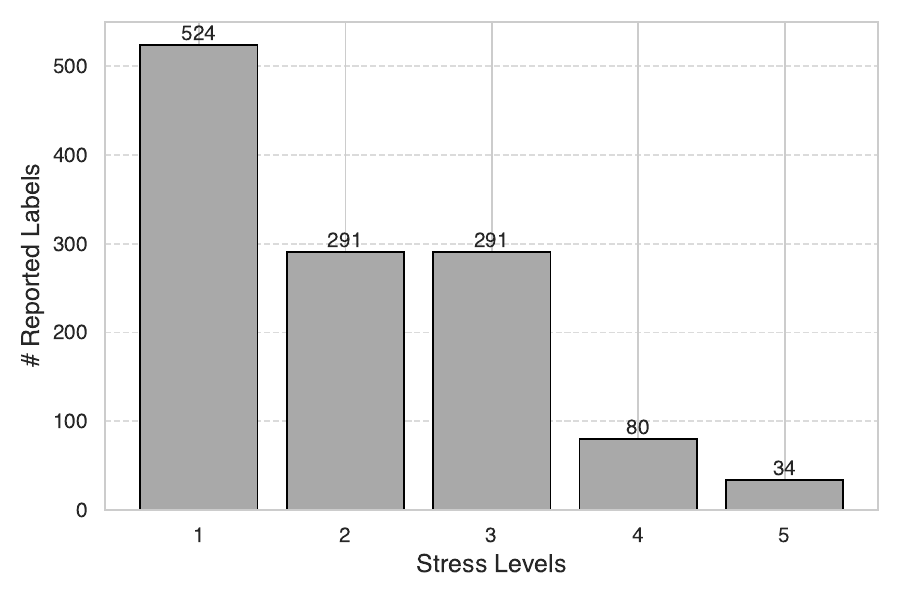}
      \vspace{-4mm}
      \caption{Distribution of Stress Labels}
      \label{fig:stress-lbl}
      \centering
\end{figure}

The EMA protocol is set to activate no more than seven times daily, prompting the participants to rate their stress levels on a five-point scale: (1) not at all, (2) a little bit, (3) some, (4) a lot, and (5) extremely. These self-reported stress levels, along with their associated timestamps, are archived in the cloud for future analysis. Each 15-minute interval of accumulated physiological and contextual data is labeled in accordance with the nearest subsequent EMA response. The distribution of these labels can be seen in Figure \ref{fig:stress-lbl}.

\subsection{Evaluation and Results}

In order to conduct a comprehensive comparison between our online context-aware active learning method for stress detection and previously offline variant, we have deliberately employed identical classification algorithms for both studies. 

Three distinct classification techniques have been used: Support Vector Machines (SVM) \cite{hearst1998support}, Random Forest \cite{breiman2001random}, and XGBoost \cite{chen2015xgboost}. SVM finds a hyperplane in high-dimensional space to separate data classes. Random Forest uses multiple decision tree classifiers on dataset subsets, improving predictive accuracy while avoiding overfitting. Additionally, XGBoost is employed, providing an effective gradient-boosted trees implementation.


The utilization of these diverse classification techniques enables a comprehensive and robust evaluation of our proposed stress detection algorithm.

Our stress detection models are classified into two categories: single-modal and multi-modal algorithms. Within the single-modal algorithm, solely the PPG signal is employed for constructing the stress detection models. On the other hand, the multi-modal algorithm utilizes both the PPG signal and contextual information (AWARE data) in the development of the proposed models.

\vspace{-0.18cm}

\subsection{Classification Performance}

\subsubsection{Experiment Detail}

In order to ensure a fair evaluation, we utilize the k-fold cross validation technique \cite{berrar2019cross} with k equal to 4.

K-fold cross-validation involves splitting the data into multiple subsets for training and testing the model. It prevents overfitting, utilizes all available data, and improves model robustness against data variations. Averaging results across folds provides a reliable way to evaluate model performance, making it valuable for model selection and hyperparameter tuning.


\subsubsection{Evaluation Metrics}


To evaluate our stress monitoring system, we use three key metrics: F1-score, precision, and recall. The F1-score assesses binary categorization test accuracy, calculated from precision and recall, where precision measures correctly identified "true positive" results and recall identifies all "true positive" results. F1-score is a weighted average of precision and recall, important for binary classification tests.

\subsubsection{Classification Performance Results}

Table \ref{tbl:perf-results} presents a comprehensive performance analysis of our novel stress detection algorithm, incorporating an online context-aware active learning approach, compared with the offline variant.

The results clearly illustrate the substantial performance enhancements achievable with the online context-aware algorithm across all evaluated metrics when compared to the offline counterpart. Notably, for the Random Forest classifier, we observe a noteworthy 11\% improvement in F1-score. The significant improvement in performance underscores the importance of employing intelligent real-time label triggering methods to identify optimal moments for sending Ecological Momentary Assessments (EMAs).

This outcome also underscores the considerable advantage of incorporating contextual awareness into our model, resulting in significant enhancements across various classification metrics and reaffirming the pivotal role of context in stress detection tasks.

\begin{table*}[!ht]
\centering
\caption{Classification Performance Results}
\label{tbl:perf-results}
\resizebox{\textwidth}{!}{%
\begin{tabular}{|c|c|c|c|c|c|c|c|c|c|c|}
\hline
            & & \multicolumn{9}{c|}{\textbf{Classification Model}} \\
            \hline
 & & \multicolumn{3}{c|}{\textbf{Random Forest}} & \multicolumn{3}{c|}{\textbf{XGBoost}} & \multicolumn{3}{c|}{\textbf{SVM}} \\
 \hline

 \textbf{Active Learning Method} & \textbf{Data} & \textbf{F1} & \textbf{Precision} & \textbf{Recall} &  \textbf{F1} & \textbf{Precision} & \textbf{Recall} & \textbf{F1} & \textbf{Precision} & \textbf{Recall}\\
\hline
Offline Context-Aware & PPG & 0.21 & 0.27 & 0.17 & 0.31 & 0.35 & 0.27 & 0.41 & 0.32 & 0.58\\
\hline
Online Context-Aware & PPG & 0.32 & 0.43 & 0.25 & 0.39 & 0.43 & 0.35 & 0.5 & 0.41 & 0.64\\
\hline
Online Context-Aware & PPG and Context & \textbf{0.36} & \textbf{0.49} & \textbf{0.28} & \textbf{0.40} & \textbf{0.45} & \textbf{0.36} & \textbf{0.52} & \textbf{0.45} & 0.61\\
\hline
\end{tabular}
}
\end{table*}

\subsection{Personalization Performance}

Leveraging the unique physiological and behavioral variations in individuals can significantly enhance the efficacy of generic models. Inspired by the potential advantages of individualized prediction models, we hypothesize that personalizing reinforcement learning models might similarly elevate data quality. To validate this premise, we initially trained a generalized representation model using the aggregated training data from all users. Subsequently, we fine-tuned this model for each user individually, aiming to discern potential enhancements in prediction accuracy. Our evaluation centered on contrasting these generalized and personalized models to elucidate the tangible benefits of our individual-based personalization strategy in data collection.

\subsubsection{Experiment Detail}

To accurately evaluate the affect of personalization in our data collection mechanism, we implemented a unique train-test splitting strategy. Our dataset comprises data from multiple users. To ensure a robust evaluation, we adopted a leave-one-subject-out cross-validation scheme. In each round of this scheme, data for each user is divided temporally into two parts. The initial half serves the purpose of model personalization, while the latter half is reserved for testing.

Two distinct models were constructed for comparative assessment:

\begin{itemize}
    \item \textbf{Plain Model}: This model is trained using the entire dataset except for the data of the user currently under consideration. For testing and evaluation, the latter half of this user's data is employed.
    \item \textbf{Personalized Model}: This model, on the other hand, is trained using the complete dataset (excluding the data of the current user) combined with the initial half of the current user's data. Again, the latter half of the user's data is utilized for testing.
\end{itemize}

Comparing these two models helps us gauge the effectiveness of our personalization strategy. By contrasting their performance, we can see how incorporating user-specific data for training improves accuracy significantly compared to using a generic global dataset.


\subsubsection{Personalization Performance Results}

Table \ref{tbl:perf-results} showcases the comparative performance of our stress detection model in both personalized and unpersonalized configurations. We present outcomes from both the ROC curve, as referenced in \ref{tbl:perf-results}, and additional performance metrics. The ROC curve assesses binary classification model efficacy by illustrating the relationship between True Positive Rate and False Positive Rate across various decision thresholds. An elevated Area Under the Curve (AUC) signifies superior model performance, underscoring its merit as a comparative measure. The findings, as depicted in the provided figure and table, reveal that adopting a personalized training approach markedly amplifies the efficacy of our stress detection strategy, as evidenced by the AUC-ROC score. Notably, when employing the XGBoost classifier, we observed a pronounced boost of approximately 10\% in the AUC-ROC score.

\begin{table*}[!ht]
\centering
\caption{Personalization Results}
\label{tbl:pers-results}
\resizebox{\textwidth}{!}{%
\begin{tabular}{|c|c|c|c|c|c|c|c|c|c|c|}
\hline
            & & \multicolumn{8}{c|}{\textbf{Classification Model}} \\
            \hline
 & & \multicolumn{2}{c|}{\textbf{Random Forest}} & \multicolumn{3}{c|}{\textbf{XGBoost}} & \multicolumn{3}{c|}{\textbf{SVM}} \\
 \hline

 \textbf{Personalized Training Method} & \textbf{F1} & \textbf{Precision} & \textbf{Recall} &  \textbf{F1} & \textbf{Precision} & \textbf{Recall} & \textbf{F1} & \textbf{Precision} & \textbf{Recall}\\
\hline
Not Personalized & 0.60 & 0.55 & 0.64 & 0.60 & 0.54 & 0.68 & 0.62 & 0.59 & 0.66\\
\hline
Personalized & \textbf{0.64} & \textbf{0.61} & \textbf{0.66} & \textbf{0.66} & \textbf{0.61} & \textbf{0.71} & 
\textbf{0.65} & \textbf{0.66} & 0.65\\
\hline
\end{tabular}
}
\end{table*}

\begin{figure*}[t]
      \includegraphics[width=1\linewidth]{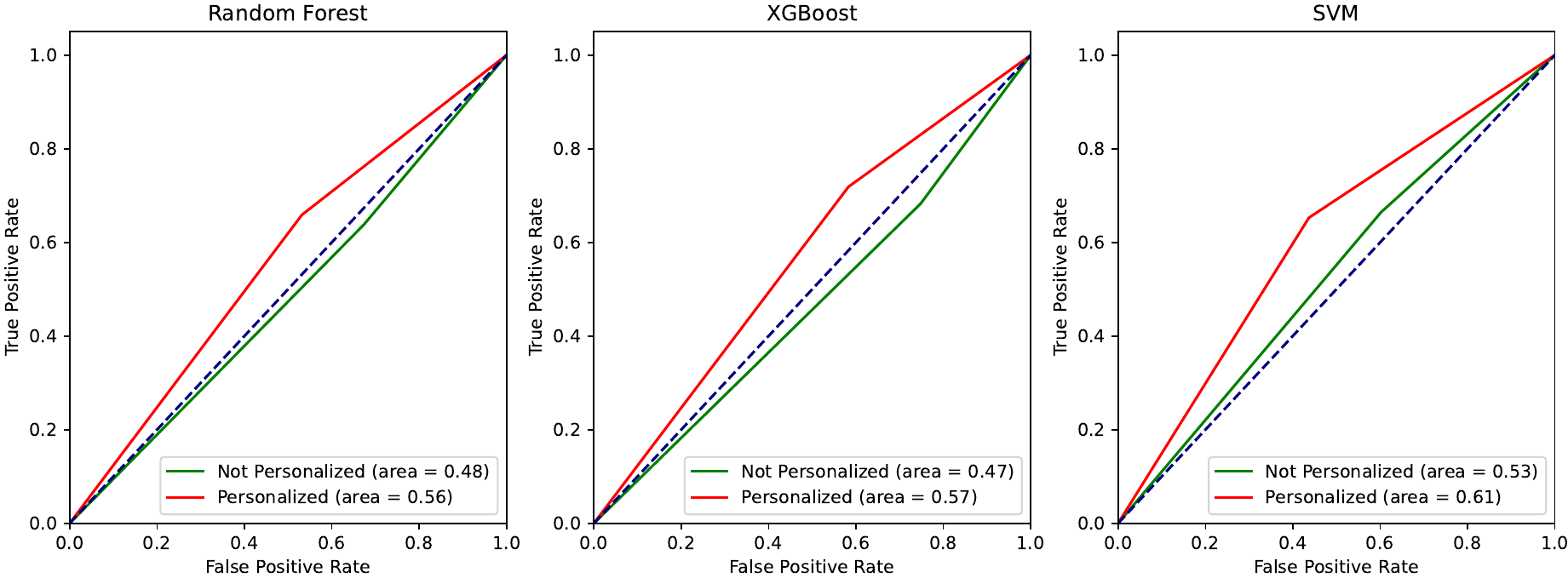}
      \vspace{-5mm}
      \caption{Presonalization ROC Curve}
      \label{fig:pers}
      \centering
      \vspace{-5mm}
\end{figure*}

\section{Conclusions}

In conclusion, this work introduced a novel contextual variant of active learning, leveraging Deep Q-Learning to incorporate individual contextual information into the decision-making process \cite{tazarv2023active, tazarv2021personalized}. In the initial phase, the implementation of a context-aware active reinforcement learning algorithm in an offline setting showcased its efficacy, resulting in a significant reduction of up to 88\% in required EMAs compared to random selection and up to 32\% compared to traditional active learning methods. Additionally, stress detection performance exhibited notable improvements, with up to a 21\% enhancement compared to random selection and up to 8\% compared to traditional active learning.

Moving to the second phase, our online implementation of the algorithm utilized active learning for EMA initiation, leveraging real-time contextual information to optimize question timings and reduce participant burden. Comparative analyses of the offline and online variants on the same dataset unequivocally demonstrated the superiority of the online algorithm, showcasing a potential improvement of up to 11\% in stress detection performance. Incorporating contextual features further improved results by 4\%, emphasizing the significance of personalization in enhancing model performance.

This study not only contributes a valuable advancement in stress detection methodologies but also underscores the pivotal role of context-awareness and online implementation in achieving superior results. The demonstrated reductions in participant burden and improvements in label accuracy signify the potential practical impact of this research in real-world applications. Future directions may explore additional personalization techniques and extend the application of context-aware active learning to diverse domains, fostering continued advancements in intelligent and user-centric systems.

\vspace{-1mm}




\bibliographystyle{ACM-Reference-Format}
\bibliography{ref} 


\begin{thebibliography}{47}


\ifx \showCODEN    \undefined \def \showCODEN     #1{\unskip}     \fi
\ifx \showDOI      \undefined \def \showDOI       #1{#1}\fi
\ifx \showISBNx    \undefined \def \showISBNx     #1{\unskip}     \fi
\ifx \showISBNxiii \undefined \def \showISBNxiii  #1{\unskip}     \fi
\ifx \showISSN     \undefined \def \showISSN      #1{\unskip}     \fi
\ifx \showLCCN     \undefined \def \showLCCN      #1{\unskip}     \fi
\ifx \shownote     \undefined \def \shownote      #1{#1}          \fi
\ifx \showarticletitle \undefined \def \showarticletitle #1{#1}   \fi
\ifx \showURL      \undefined \def \showURL       {\relax}        \fi
\providecommand\bibfield[2]{#2}
\providecommand\bibinfo[2]{#2}
\providecommand\natexlab[1]{#1}
\providecommand\showeprint[2][]{arXiv:#2}

\bibitem[ame(2021)]%
        {americanstress}
 \bibinfo{year}{2021}\natexlab{}.
\newblock \bibinfo{title}{The American Institute of Stresss}.
\newblock
\newblock
\urldef\tempurl%
\url{https://www.stress.org}
\showURL{%
\tempurl}


\bibitem[awa(2023)]%
        {awareframework}
 \bibinfo{year}{2023}\natexlab{}.
\newblock \bibinfo{title}{AWARE Framework}.
\newblock
\newblock
\urldef\tempurl%
\url{https://awareframework.com}
\showURL{%
\tempurl}


\bibitem[may(2023)]%
        {mayoclinic}
 \bibinfo{year}{2023}\natexlab{}.
\newblock \bibinfo{title}{Mayo Clinic}.
\newblock
\newblock
\urldef\tempurl%
\url{https://www.mayoclinic.org}
\showURL{%
\tempurl}


\bibitem[Alikhani et~al\mbox{.}(2023)]%
        {alikhani2023dynafuse}
\bibfield{author}{\bibinfo{person}{Hamidreza Alikhani}, \bibinfo{person}{Anil Kanduri}, \bibinfo{person}{Pasi Liljeberg}, \bibinfo{person}{Amir~M Rahmani}, {and} \bibinfo{person}{Nikil Dutt}.} \bibinfo{year}{2023}\natexlab{}.
\newblock \showarticletitle{DynaFuse: Dynamic Fusion for Resource Efficient Multi-Modal Machine Learning Inference}.
\newblock \bibinfo{journal}{\emph{IEEE Embedded Systems Letters}} (\bibinfo{year}{2023}).
\newblock


\bibitem[Alikhani et~al\mbox{.}(2024)]%
        {alikhani2024seal}
\bibfield{author}{\bibinfo{person}{Hamidreza Alikhani}, \bibinfo{person}{Ziyu Wang}, \bibinfo{person}{Anil Kanduri}, \bibinfo{person}{Pasi Lilieberg}, \bibinfo{person}{Amir~M Rahmani}, {and} \bibinfo{person}{Nikil Dutt}.} \bibinfo{year}{2024}\natexlab{}.
\newblock \showarticletitle{SEAL: Sensing Efficient Active Learning on Wearables through Context-awareness}. In \bibinfo{booktitle}{\emph{2024 Design, Automation \& Test in Europe Conference \& Exhibition (DATE)}}. IEEE, \bibinfo{pages}{1--2}.
\newblock


\bibitem[Allen(2007)]%
        {allen2007photoplethysmography}
\bibfield{author}{\bibinfo{person}{John Allen}.} \bibinfo{year}{2007}\natexlab{}.
\newblock \showarticletitle{Photoplethysmography and its application in clinical physiological measurement}.
\newblock \bibinfo{journal}{\emph{Physiological measurement}} \bibinfo{volume}{28}, \bibinfo{number}{3} (\bibinfo{year}{2007}), \bibinfo{pages}{R1}.
\newblock


\bibitem[Battalio et~al\mbox{.}(2021)]%
        {battalio2021sense2stop}
\bibfield{author}{\bibinfo{person}{Samuel~L Battalio} {et~al\mbox{.}}} \bibinfo{year}{2021}\natexlab{}.
\newblock \showarticletitle{Sense2Stop: a micro-randomized trial using wearable sensors to optimize a just-in-time-adaptive stress management intervention for smoking relapse prevention}.
\newblock \bibinfo{journal}{\emph{Contemporary Clinical Trials}}  \bibinfo{volume}{109} (\bibinfo{year}{2021}), \bibinfo{pages}{106534}.
\newblock


\bibitem[Berrar et~al\mbox{.}(2019)]%
        {berrar2019cross}
\bibfield{author}{\bibinfo{person}{Daniel Berrar} {et~al\mbox{.}}} \bibinfo{year}{2019}\natexlab{}.
\newblock \bibinfo{title}{Cross-Validation.}
\newblock
\newblock


\bibitem[Breiman(2001)]%
        {breiman2001random}
\bibfield{author}{\bibinfo{person}{Leo Breiman}.} \bibinfo{year}{2001}\natexlab{}.
\newblock \showarticletitle{Random forests}.
\newblock \bibinfo{journal}{\emph{Machine learning}}  \bibinfo{volume}{45} (\bibinfo{year}{2001}), \bibinfo{pages}{5--32}.
\newblock


\bibitem[Burke et~al\mbox{.}(2017)]%
        {burke2017ecological}
\bibfield{author}{\bibinfo{person}{Lora~E Burke}, \bibinfo{person}{Saul Shiffman}, \bibinfo{person}{Edvin Music}, \bibinfo{person}{Mindi~A Styn}, \bibinfo{person}{Andrea Kriska}, \bibinfo{person}{Asim Smailagic}, \bibinfo{person}{Daniel Siewiorek}, \bibinfo{person}{Linda~J Ewing}, \bibinfo{person}{Eileen Chasens}, \bibinfo{person}{Brian French}, {et~al\mbox{.}}} \bibinfo{year}{2017}\natexlab{}.
\newblock \showarticletitle{Ecological momentary assessment in behavioral research: addressing technological and human participant challenges}.
\newblock \bibinfo{journal}{\emph{Journal of medical Internet research}} \bibinfo{volume}{19}, \bibinfo{number}{3} (\bibinfo{year}{2017}), \bibinfo{pages}{e77}.
\newblock


\bibitem[Can et~al\mbox{.}(2020)]%
        {can2020real}
\bibfield{author}{\bibinfo{person}{Yekta~Said Can} {et~al\mbox{.}}} \bibinfo{year}{2020}\natexlab{}.
\newblock \showarticletitle{Real-life stress level monitoring using smart bands in the light of contextual information}.
\newblock \bibinfo{journal}{\emph{IEEE Sensors Journal}} \bibinfo{volume}{20}, \bibinfo{number}{15} (\bibinfo{year}{2020}), \bibinfo{pages}{8721--8730}.
\newblock


\bibitem[Castaneda et~al\mbox{.}(2018)]%
        {castaneda2018review}
\bibfield{author}{\bibinfo{person}{Denisse Castaneda} {et~al\mbox{.}}} \bibinfo{year}{2018}\natexlab{}.
\newblock \showarticletitle{A review on wearable photoplethysmography sensors and their potential future applications in health care}.
\newblock \bibinfo{journal}{\emph{International journal of biosensors \& bioelectronics}} \bibinfo{volume}{4}, \bibinfo{number}{4} (\bibinfo{year}{2018}), \bibinfo{pages}{195}.
\newblock


\bibitem[Charlton et~al\mbox{.}(2018)]%
        {charlton2018assessing}
\bibfield{author}{\bibinfo{person}{Peter~H Charlton} {et~al\mbox{.}}} \bibinfo{year}{2018}\natexlab{}.
\newblock \showarticletitle{Assessing mental stress from the photoplethysmogram: a numerical study}.
\newblock \bibinfo{journal}{\emph{Physiological measurement}} \bibinfo{volume}{39}, \bibinfo{number}{5} (\bibinfo{year}{2018}), \bibinfo{pages}{054001}.
\newblock


\bibitem[Chen et~al\mbox{.}(2015)]%
        {chen2015xgboost}
\bibfield{author}{\bibinfo{person}{Tianqi Chen} {et~al\mbox{.}}} \bibinfo{year}{2015}\natexlab{}.
\newblock \showarticletitle{Xgboost: extreme gradient boosting}.
\newblock \bibinfo{journal}{\emph{R package version 0.4-2}} \bibinfo{volume}{1}, \bibinfo{number}{4} (\bibinfo{year}{2015}), \bibinfo{pages}{1--4}.
\newblock


\bibitem[Cheng et~al\mbox{.}(2024a)]%
        {cheng2024vetrass}
\bibfield{author}{\bibinfo{person}{Ming Cheng}, \bibinfo{person}{Bowen Zhang}, \bibinfo{person}{Ziyu Wang}, \bibinfo{person}{Ziyi Zhou}, \bibinfo{person}{Weiqi Feng}, \bibinfo{person}{Yi Lyu}, {and} \bibinfo{person}{Xingjian Diao}.} \bibinfo{year}{2024}\natexlab{a}.
\newblock \showarticletitle{VeTraSS: Vehicle Trajectory Similarity Search Through Graph Modeling and Representation Learning}.
\newblock \bibinfo{journal}{\emph{arXiv preprint arXiv:2404.08021}} (\bibinfo{year}{2024}).
\newblock


\bibitem[Cheng et~al\mbox{.}(2024b)]%
        {cheng2024efflex}
\bibfield{author}{\bibinfo{person}{Ming Cheng}, \bibinfo{person}{Ziyi Zhou}, \bibinfo{person}{Bowen Zhang}, \bibinfo{person}{Ziyu Wang}, \bibinfo{person}{Jiaqi Gan}, \bibinfo{person}{Ziang Ren}, \bibinfo{person}{Weiqi Feng}, \bibinfo{person}{Yi Lyu}, \bibinfo{person}{Hefan Zhang}, {and} \bibinfo{person}{Xingjian Diao}.} \bibinfo{year}{2024}\natexlab{b}.
\newblock \showarticletitle{Efflex: Efficient and Flexible Pipeline for Spatio-Temporal Trajectory Graph Modeling and Representation Learning}. In \bibinfo{booktitle}{\emph{Proceedings of the IEEE/CVF Conference on Computer Vision and Pattern Recognition}}. \bibinfo{pages}{2546--2555}.
\newblock


\bibitem[Cohen et~al\mbox{.}(1994)]%
        {cohen1994perceived}
\bibfield{author}{\bibinfo{person}{Sheldon Cohen}, \bibinfo{person}{Tom Kamarck}, \bibinfo{person}{Robin Mermelstein}, {et~al\mbox{.}}} \bibinfo{year}{1994}\natexlab{}.
\newblock \showarticletitle{Perceived stress scale}.
\newblock \bibinfo{journal}{\emph{Measuring stress: A guide for health and social scientists}} \bibinfo{volume}{10}, \bibinfo{number}{2} (\bibinfo{year}{1994}), \bibinfo{pages}{1--2}.
\newblock


\bibitem[Dogan et~al\mbox{.}(2022)]%
        {dogan2022stress}
\bibfield{author}{\bibinfo{person}{Gulin Dogan} {et~al\mbox{.}}} \bibinfo{year}{2022}\natexlab{}.
\newblock \showarticletitle{Stress detection using experience sampling: A systematic mapping study}.
\newblock \bibinfo{journal}{\emph{International Journal of Environmental Research and Public Health}} \bibinfo{volume}{19}, \bibinfo{number}{9} (\bibinfo{year}{2022}), \bibinfo{pages}{5693}.
\newblock


\bibitem[Fahrenberg et~al\mbox{.}(2007)]%
        {fahrenberg2007ambulatory}
\bibfield{author}{\bibinfo{person}{Jochen Fahrenberg}, \bibinfo{person}{Michael Myrtek}, \bibinfo{person}{Kurt Pawlik}, {and} \bibinfo{person}{Meinrad Perrez}.} \bibinfo{year}{2007}\natexlab{}.
\newblock \showarticletitle{Ambulatory assessment-monitoring behavior in daily life settings}.
\newblock \bibinfo{journal}{\emph{European Journal of Psychological Assessment}} \bibinfo{volume}{23}, \bibinfo{number}{4} (\bibinfo{year}{2007}), \bibinfo{pages}{206--213}.
\newblock


\bibitem[Giannakakis et~al\mbox{.}(2019)]%
        {giannakakis2019review}
\bibfield{author}{\bibinfo{person}{Giorgos Giannakakis}, \bibinfo{person}{Dimitris Grigoriadis}, \bibinfo{person}{Katerina Giannakaki}, \bibinfo{person}{Olympia Simantiraki}, \bibinfo{person}{Alexandros Roniotis}, {and} \bibinfo{person}{Manolis Tsiknakis}.} \bibinfo{year}{2019}\natexlab{}.
\newblock \showarticletitle{Review on psychological stress detection using biosignals}.
\newblock \bibinfo{journal}{\emph{IEEE Transactions on Affective Computing}} \bibinfo{volume}{13}, \bibinfo{number}{1} (\bibinfo{year}{2019}), \bibinfo{pages}{440--460}.
\newblock


\bibitem[Han et~al\mbox{.}(2020)]%
        {han2020objective}
\bibfield{author}{\bibinfo{person}{Hee~Jeong Han} {et~al\mbox{.}}} \bibinfo{year}{2020}\natexlab{}.
\newblock \showarticletitle{Objective stress monitoring based on wearable sensors in everyday settings}.
\newblock \bibinfo{journal}{\emph{Journal of Medical Engineering \& Technology}} \bibinfo{volume}{44}, \bibinfo{number}{4} (\bibinfo{year}{2020}), \bibinfo{pages}{177--189}.
\newblock


\bibitem[Hearst et~al\mbox{.}(1998)]%
        {hearst1998support}
\bibfield{author}{\bibinfo{person}{Marti~A. Hearst} {et~al\mbox{.}}} \bibinfo{year}{1998}\natexlab{}.
\newblock \showarticletitle{Support vector machines}.
\newblock \bibinfo{journal}{\emph{IEEE Intelligent Systems and their applications}} \bibinfo{volume}{13}, \bibinfo{number}{4} (\bibinfo{year}{1998}), \bibinfo{pages}{18--28}.
\newblock


\bibitem[Hester et~al\mbox{.}(2018)]%
        {hester2018deep}
\bibfield{author}{\bibinfo{person}{Todd Hester} {et~al\mbox{.}}} \bibinfo{year}{2018}\natexlab{}.
\newblock \showarticletitle{Deep q-learning from demonstrations}. In \bibinfo{booktitle}{\emph{Proceedings of the AAAI conference on artificial intelligence}}, Vol.~\bibinfo{volume}{32.1}.
\newblock


\bibitem[Holmes and Rahe(1967)]%
        {holmes1967social}
\bibfield{author}{\bibinfo{person}{Thomas~H Holmes} {and} \bibinfo{person}{Richard~H Rahe}.} \bibinfo{year}{1967}\natexlab{}.
\newblock \showarticletitle{The social readjustment rating scale.}
\newblock \bibinfo{journal}{\emph{Journal of psychosomatic research}} (\bibinfo{year}{1967}).
\newblock


\bibitem[Kanduri et~al\mbox{.}(2023)]%
        {kanduri2023edge}
\bibfield{author}{\bibinfo{person}{Anil Kanduri}, \bibinfo{person}{Sina Shahhosseini}, \bibinfo{person}{Emad~Kasaeyan Naeini}, \bibinfo{person}{Hamidreza Alikhani}, \bibinfo{person}{Pasi Liljeberg}, \bibinfo{person}{Nikil Dutt}, {and} \bibinfo{person}{Amir~M Rahmani}.} \bibinfo{year}{2023}\natexlab{}.
\newblock \showarticletitle{Edge-centric Optimization of Multi-modal ML-driven eHealth Applications}.
\newblock In \bibinfo{booktitle}{\emph{Embedded Machine Learning for Cyber-Physical, IoT, and Edge Computing: Use Cases and Emerging Challenges}}. \bibinfo{publisher}{Springer}, \bibinfo{pages}{95--125}.
\newblock


\bibitem[Kotsiantis et~al\mbox{.}(2007)]%
        {kotsiantis2007supervised}
\bibfield{author}{\bibinfo{person}{Sotiris~B Kotsiantis} {et~al\mbox{.}}} \bibinfo{year}{2007}\natexlab{}.
\newblock \showarticletitle{Supervised machine learning: A review of classification techniques}.
\newblock \bibinfo{journal}{\emph{Emerging artificial intelligence applications in computer engineering}} \bibinfo{volume}{160}, \bibinfo{number}{1} (\bibinfo{year}{2007}), \bibinfo{pages}{3--24}.
\newblock


\bibitem[Larradet et~al\mbox{.}(2020)]%
        {larradet2020toward}
\bibfield{author}{\bibinfo{person}{Fanny Larradet} {et~al\mbox{.}}} \bibinfo{year}{2020}\natexlab{}.
\newblock \showarticletitle{Toward emotion recognition from physiological signals in the wild: approaching the methodological issues in real-life data collection}.
\newblock \bibinfo{journal}{\emph{Frontiers in psychology}}  \bibinfo{volume}{11} (\bibinfo{year}{2020}), \bibinfo{pages}{1111}.
\newblock


\bibitem[Mundnich et~al\mbox{.}(2020)]%
        {mundnich2020tiles}
\bibfield{author}{\bibinfo{person}{Karel Mundnich} {et~al\mbox{.}}} \bibinfo{year}{2020}\natexlab{}.
\newblock \showarticletitle{TILES-2018, a longitudinal physiologic and behavioral data set of hospital workers}.
\newblock \bibinfo{journal}{\emph{Scientific Data}} \bibinfo{volume}{7}, \bibinfo{number}{1} (\bibinfo{year}{2020}), \bibinfo{pages}{354}.
\newblock


\bibitem[Plappert(2016)]%
        {plappert2016kerasrl}
\bibfield{author}{\bibinfo{person}{Matthias Plappert}.} \bibinfo{year}{2016}\natexlab{}.
\newblock \bibinfo{title}{keras-rl}.
\newblock \bibinfo{howpublished}{\url{https://github.com/keras-rl/keras-rl}}.
\newblock


\bibitem[Sannino et~al\mbox{.}(2014)]%
        {sannino2014mobile}
\bibfield{author}{\bibinfo{person}{Giovanna Sannino} {et~al\mbox{.}}} \bibinfo{year}{2014}\natexlab{}.
\newblock \showarticletitle{A mobile system for real-time context-aware monitoring of patients’ health and fainting}.
\newblock \bibinfo{journal}{\emph{International journal of data mining and bioinformatics}} \bibinfo{volume}{10}, \bibinfo{number}{4} (\bibinfo{year}{2014}), \bibinfo{pages}{407--423}.
\newblock


\bibitem[Sarhaddi et~al\mbox{.}(2022)]%
        {sarhaddi2022comprehensive}
\bibfield{author}{\bibinfo{person}{Fatemeh Sarhaddi} {et~al\mbox{.}}} \bibinfo{year}{2022}\natexlab{}.
\newblock \showarticletitle{A comprehensive accuracy assessment of Samsung smartwatch heart rate and heart rate variability}.
\newblock \bibinfo{journal}{\emph{PloS one}} \bibinfo{volume}{17}, \bibinfo{number}{12} (\bibinfo{year}{2022}), \bibinfo{pages}{e0268361}.
\newblock


\bibitem[Seok et~al\mbox{.}(2021)]%
        {seok2021motion}
\bibfield{author}{\bibinfo{person}{Dongyeol Seok} {et~al\mbox{.}}} \bibinfo{year}{2021}\natexlab{}.
\newblock \showarticletitle{Motion artifact removal techniques for wearable EEG and PPG sensor systems}.
\newblock \bibinfo{journal}{\emph{Frontiers in Electronics}}  \bibinfo{volume}{2} (\bibinfo{year}{2021}), \bibinfo{pages}{685513}.
\newblock


\bibitem[Settles(2009)]%
        {settles2009active}
\bibfield{author}{\bibinfo{person}{Burr Settles}.} \bibinfo{year}{2009}\natexlab{}.
\newblock \showarticletitle{Active learning literature survey}.
\newblock  (\bibinfo{year}{2009}).
\newblock


\bibitem[Sina et~al\mbox{.}(2023)]%
        {zotcare}
\bibfield{author}{\bibinfo{person}{Labbaf Sina} {et~al\mbox{.}}} \bibinfo{year}{2023}\natexlab{}.
\newblock \showarticletitle{ZotCare: a flexible, personalizable, and affordable mhealth service provider}.
\newblock \bibinfo{journal}{\emph{Front. Digit. Health}} (\bibinfo{year}{2023}).
\newblock


\bibitem[Steptoe et~al\mbox{.}(2003)]%
        {steptoe2003socioeconomic}
\bibfield{author}{\bibinfo{person}{Andrew Steptoe}, \bibinfo{person}{Sabine Kunz-Ebrecht}, \bibinfo{person}{Natalie Owen}, \bibinfo{person}{Pamela~J Feldman}, \bibinfo{person}{Gonneke Willemsen}, \bibinfo{person}{Clemens Kirschbaum}, {and} \bibinfo{person}{Michael Marmot}.} \bibinfo{year}{2003}\natexlab{}.
\newblock \showarticletitle{Socioeconomic status and stress-related biological responses over the working day}.
\newblock \bibinfo{journal}{\emph{Psychosomatic medicine}} \bibinfo{volume}{65}, \bibinfo{number}{3} (\bibinfo{year}{2003}), \bibinfo{pages}{461--470}.
\newblock


\bibitem[Stojchevska et~al\mbox{.}(2022)]%
        {stojchevska2022assessing}
\bibfield{author}{\bibinfo{person}{Marija Stojchevska} {et~al\mbox{.}}} \bibinfo{year}{2022}\natexlab{}.
\newblock \showarticletitle{Assessing the added value of context during stress detection from wearable data}.
\newblock \bibinfo{journal}{\emph{BMC Medical Informatics and Decision Making}} \bibinfo{volume}{22}, \bibinfo{number}{1} (\bibinfo{year}{2022}), \bibinfo{pages}{268}.
\newblock


\bibitem[Tazarv et~al\mbox{.}(2021)]%
        {tazarv2021personalized}
\bibfield{author}{\bibinfo{person}{Ali Tazarv} {et~al\mbox{.}}} \bibinfo{year}{2021}\natexlab{}.
\newblock \showarticletitle{Personalized stress monitoring using wearable sensors in everyday settings}. In \bibinfo{booktitle}{\emph{2021 43rd Annual International Conference of the IEEE Engineering in Medicine \& Biology Society (EMBC)}}. IEEE, \bibinfo{pages}{7332--7335}.
\newblock


\bibitem[Tazarv et~al\mbox{.}(2023)]%
        {tazarv2023active}
\bibfield{author}{\bibinfo{person}{Ali Tazarv} {et~al\mbox{.}}} \bibinfo{year}{2023}\natexlab{}.
\newblock \showarticletitle{Active Reinforcement Learning for Personalized Stress Monitoring in Everyday Settings}.
\newblock \bibinfo{journal}{\emph{arXiv preprint arXiv:2305.00111}} (\bibinfo{year}{2023}).
\newblock


\bibitem[Van~Gent et~al\mbox{.}(2019)]%
        {van2019heartpy}
\bibfield{author}{\bibinfo{person}{Paul Van~Gent} {et~al\mbox{.}}} \bibinfo{year}{2019}\natexlab{}.
\newblock \showarticletitle{HeartPy: A novel heart rate algorithm for the analysis of noisy signals}.
\newblock \bibinfo{journal}{\emph{Transportation research part F: traffic psychology and behaviour}}  \bibinfo{volume}{66} (\bibinfo{year}{2019}), \bibinfo{pages}{368--378}.
\newblock


\bibitem[Vashisht et~al\mbox{.}(2014)]%
        {vashisht2014study}
\bibfield{author}{\bibinfo{person}{Geetika Vashisht} {et~al\mbox{.}}} \bibinfo{year}{2014}\natexlab{}.
\newblock \showarticletitle{A study on the Tizen Operating System}.
\newblock \bibinfo{journal}{\emph{International Journal of Computer Trends and Technology}} \bibinfo{volume}{12}, \bibinfo{number}{1} (\bibinfo{year}{2014}), \bibinfo{pages}{14--15}.
\newblock


\bibitem[Wang et~al\mbox{.}(2020a)]%
        {wang2020social}
\bibfield{author}{\bibinfo{person}{Weichen Wang} {et~al\mbox{.}}} \bibinfo{year}{2020}\natexlab{a}.
\newblock \showarticletitle{Social sensing: assessing social functioning of patients living with schizophrenia using mobile phone sensing}. In \bibinfo{booktitle}{\emph{Proceedings of the 2020 CHI conference on human factors in computing systems}}. \bibinfo{pages}{1--15}.
\newblock


\bibitem[Wang et~al\mbox{.}(2020b)]%
        {wang2020guardhealth}
\bibfield{author}{\bibinfo{person}{Ziyu Wang} {et~al\mbox{.}}} \bibinfo{year}{2020}\natexlab{b}.
\newblock \showarticletitle{GuardHealth: Blockchain empowered secure data management and Graph Convolutional Network enabled anomaly detection in smart healthcare}.
\newblock \bibinfo{journal}{\emph{J. Parallel and Distrib. Comput.}}  \bibinfo{volume}{142} (\bibinfo{year}{2020}), \bibinfo{pages}{1--12}.
\newblock


\bibitem[Wang et~al\mbox{.}(2024)]%
        {wang2024differential}
\bibfield{author}{\bibinfo{person}{Ziyu Wang}, \bibinfo{person}{Zhongqi Yang}, \bibinfo{person}{Iman Azimi}, {and} \bibinfo{person}{Amir~M Rahmani}.} \bibinfo{year}{2024}\natexlab{}.
\newblock \showarticletitle{Differential private federated transfer learning for mental health monitoring in everyday settings: A case study on stress detection}.
\newblock \bibinfo{journal}{\emph{arXiv preprint arXiv:2402.10862}} (\bibinfo{year}{2024}).
\newblock


\bibitem[Yang et~al\mbox{.}(2022)]%
        {yang2022zebra}
\bibfield{author}{\bibinfo{person}{Xinyu Yang}, \bibinfo{person}{Haoyuan Liu}, \bibinfo{person}{Ziyu Wang}, {and} \bibinfo{person}{Peng Gao}.} \bibinfo{year}{2022}\natexlab{}.
\newblock \showarticletitle{Zebra: Deeply integrating system-level provenance search and tracking for efficient attack investigation}.
\newblock \bibinfo{journal}{\emph{arXiv preprint arXiv:2211.05403}} (\bibinfo{year}{2022}).
\newblock


\bibitem[Yao et~al\mbox{.}(2020)]%
        {yao2020privacy}
\bibfield{author}{\bibinfo{person}{Yuanfan Yao} {et~al\mbox{.}}} \bibinfo{year}{2020}\natexlab{}.
\newblock \showarticletitle{Privacy-preserving and energy efficient task offloading for collaborative mobile computing in IoT: An ADMM approach}.
\newblock \bibinfo{journal}{\emph{Computers \& Security}}  \bibinfo{volume}{96} (\bibinfo{year}{2020}), \bibinfo{pages}{101886}.
\newblock


\bibitem[Yu et~al\mbox{.}(2020)]%
        {yu2020passive}
\bibfield{author}{\bibinfo{person}{H Yu} {et~al\mbox{.}}} \bibinfo{year}{2020}\natexlab{}.
\newblock \showarticletitle{Passive sensor data based future mood health and stress prediction: User adaptation using deep learning; passive sensor data based future mood health and stress prediction: User adaptation using deep learning}.
\newblock  (\bibinfo{year}{2020}).
\newblock


\bibitem[Yu et~al\mbox{.}(2022)]%
        {yu2022semi}
\bibfield{author}{\bibinfo{person}{Han Yu} {et~al\mbox{.}}} \bibinfo{year}{2022}\natexlab{}.
\newblock \showarticletitle{Semi-supervised learning and data augmentation in wearable-based momentary stress detection in the wild}.
\newblock \bibinfo{journal}{\emph{arXiv preprint arXiv:2202.12935}} (\bibinfo{year}{2022}).
\newblock


\end{thebibliography}

\end{document}